\documentclass[letterpaper]{article} 
\usepackage{paper} 
\nocopyright
\usepackage{times}  
\usepackage{helvet} 
\usepackage{courier} 
\usepackage[hyphens]{url}  
\usepackage{graphicx} 
\usepackage{natbib} 
\usepackage{caption} 
\usepackage{algorithm}
\usepackage{algorithmic}
\usepackage{multirow}
\usepackage{amsmath}
\usepackage{xcolor}
\usepackage{enumitem}
\usepackage{amssymb}
\usepackage{adjustbox}
\usepackage{algorithm}
\usepackage{algorithmic}
\usepackage{ragged2e}
\usepackage{multirow}
\usepackage{makecell}

\usepackage{newfloat}
\usepackage{listings}
\DeclareCaptionStyle{ruled}{labelfont=normalfont,labelsep=colon,strut=off} 
\floatstyle{ruled}
\newfloat{listing}{tb}{lst}{}
\floatname{listing}{Listing}
\title{Interactive DualChecker for Mitigating Hallucinations in Distilling Large Language Models}
\author {
    Meiyun Wang\textsuperscript{\rm 1}\thanks{Corresponding Author},
    Masahiro Suzuki\textsuperscript{\rm 1},
    Hiroki Sakaji\textsuperscript{\rm 2},
    Kiyoshi Izumi \textsuperscript{\rm 1},
}
\affiliations {
    \textsuperscript{\rm 1}The University of Tokyo\\
    \textsuperscript{\rm 2}Hokkaido University\\
    omiun20@g.ecc.u-tokyo.ac.jp, msuzuki@g.ecc.u-tokyo.ac.jp, sakaji@ist.hokudai.ac.jp, izumi@sys.t.u-tokyo.ac.jp
}
\begin{document}

\maketitle

\begin{abstract}
Large Language Models (LLMs) have demonstrated exceptional capabilities across various machine learning (ML) tasks. Given the high costs of creating annotated datasets for supervised learning, LLMs offer a valuable alternative by enabling effective few-shot in-context learning. However, these models can produce hallucinations, particularly in domains with incomplete knowledge. Additionally, current methods for knowledge distillation using LLMs often struggle to enhance the effectiveness of both teacher and student models. To address these challenges, we introduce \texttt{DualChecker}, an innovative framework designed to mitigate hallucinations and improve the performance of both teacher and student models during knowledge distillation. \texttt{DualChecker} employs ContextAligner to ensure that the context provided by teacher models aligns with human labeling standards. It also features a dynamic checker system that enhances model interaction: one component re-prompts teacher models with more detailed content when they show low confidence, and another identifies borderline cases from student models to refine the teaching templates. This interactive process promotes continuous improvement and effective knowledge transfer between the models. We evaluate \texttt{DualChecker} using a green innovation textual dataset that includes binary, multiclass, and token classification tasks. The experimental results show that \texttt{DualChecker} significantly outperforms existing state-of-the-art methods, achieving up to a 17\% improvement in F1 score for teacher models and 10\% for student models. Notably, student models fine-tuned with LLM predictions perform comparably to those fine-tuned with actual data, even in a challenging domain. We make all datasets, models, and code from this research publicly available.\footnote{\url{https://github.com/Kirawang23/DualChecker}}
\end{abstract}

\section{Introduction}

The advent of Large Language Models (LLMs) has revolutionized artificial intelligence, providing comprehensive end-to-end solutions for numerous machine learning (ML) tasks \cite{chang2024survey, huang2024position}. Traditional ML approaches predominantly rely on supervised learning, which necessitates large annotated datasets to achieve high performance. In contrast, LLMs, trained on trillions of tokens and possessing hundreds of billions of parameters, can function as extensive knowledge bases and excel in various tasks through in-context learning, without requiring additional training \cite{kojima2022large, brown2020language}.

\begin{figure}[t]
\centering
\includegraphics[width=\linewidth]{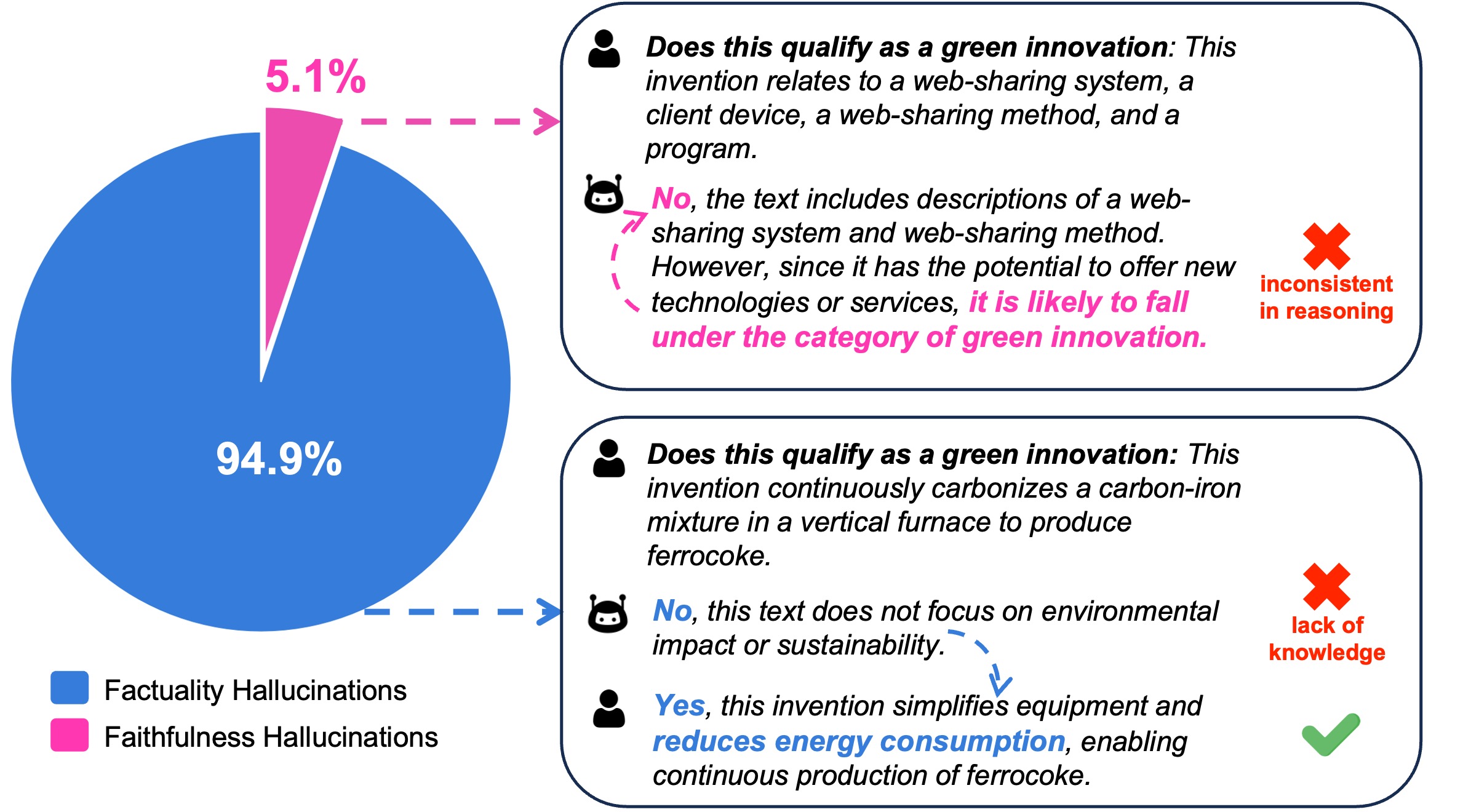}
\vspace{-0.08in}
\caption{Distribution of Hallucination Types in a Preliminary Experiment.}
\label{fig:intro}
\vspace{-0.17in}
\end{figure}

However, LLMs are prone to hallucination issues, manifesting as either factual inaccuracies or inconsistencies in responses, known respectively as factuality and faithfulness hallucinations \cite{huang2023survey}. Figure \ref{fig:intro} shows the distribution of hallucination types observed in a preliminary experiment involving 200 labeled samples generated by GPT-3.5 Turbo \footnote{\url{https://platform.openai.com/docs/models}} using zero-shot prompting from our green innovation dataset. The results indicate a predominant issue of factuality hallucinations at 94.9\%, compared to faithfulness hallucinations at only 5.1\% in domain adaptation.

Existing methods use external knowledge to supplement missing information for factuality hallucinations \cite{yao2023react, ram2023context} and focus on mitigating language model overconfidence to improve faithfulness \cite{chen2022towards, schuster2022confident, zhao2023pareto}. However, these solutions face significant challenges: a) constructing external knowledge bases is expensive, and input length limits make it difficult to determine how much external knowledge to feed into the model; b) domain adaptation is challenging due to the variability and complexity of human labeling standards; c) few studies address both factuality and faithfulness hallucinations; and d) most methods require costly additional pre-training or fine-tuning to achieve high accuracy.

To address these limitations, we introduce \texttt{DualChecker}, a novel interactive framework that mitigates hallucinations and improves the performance of both teacher and student models in knowledge distillation. \texttt{DualChecker} uses ContextAligner to align model outputs with human labeling standards by retrieving similar data from the training set and generating explanations that match human logic. Additionally, the interactive checker system collects confidence scores from LLM responses, re-prompting the model with more details when confidence is low to ensure consistency. It also identifies borderline cases from the student model, generates rationales for these cases, and incorporates them into the teaching templates to improve performance.

We conduct experiments within the challenging green innovation domain, tackling tasks such as binary green innovation classification, multiclass path classification, and token-level causality classification. The experimental results highlight the superior performance of \texttt{DualChecker} compared to existing state-of-the-art methods across all tasks. Notably, \texttt{DualChecker} achieves up to a \textbf{17\%} improvement in F1 score for teacher models and a \textbf{10\%} improvement for student models. Our backbone models include both the black-box GPT-3.5 Turbo and the white-box Llama 2 \cite{touvron2023llama}. Remarkably, \texttt{DualChecker} excels with both models, and student models fine-tuned using our predictions show performance on par with those fine-tuned with actual data.

Our contributions are summarized as follows:
\begin{itemize}[noitemsep,leftmargin=*,align=left]
\item \textbf{Introduction of DualChecker:} We propose a novel framework, \texttt{DualChecker}, which leverages LLMs for knowledge distillation. This framework provides a robust solution for many challenging tasks where LLMs lack sufficient knowledge.
\item \textbf{Mitigation of Hallucinations:} \texttt{DualChecker} effectively addresses hallucinations by employing ContextAligner and an interactive checker system to ensure accurate and reliable outputs.
\item \textbf{Improvement in Knowledge Distillation:} Experimental results demonstrate that our approach significantly boosts the performance of both teacher and student models in the knowledge distillation process.
\item \textbf{Open Source Contributions:} We are committed to advancing the field through transparency and collaboration. Therefore, we publicly release all datasets, models, and code from this research.
\end{itemize}

\vspace{-0.08in}
\section{Related Work}
\subsection{Hallucination Mitigation}
\vspace{-0.03in}
Hallucinations in language models (LMs) often arise from the optimization techniques used during training. Methods like Maximum Likelihood Estimation (MLE) with teacher forcing can cause models to replicate training data without genuine understanding, leading to hallucinations during inference \cite{kang2020improved}. To address this issue, researchers have proposed various solutions for different types of hallucinations in LLMs.

Improving the knowledge LLMs lack during reasoning has shown promise for factuality hallucinations. For instance, \cite{sun2023contrastive} employs contrastive learning to minimize the influence of confusing negative knowledge in conversations. Similarly, \cite{sahu2023promptmix} generates challenging text near class boundaries to diversify and strengthen the training data. Additionally, some studies leverage knowledge graphs as external sources to enhance reasoning capabilities \cite{guan2024mitigating, shi2023hallucination}.

However, these methods face limitations due to LLMs' input length constraints, which can make external knowledge insufficient for a comprehensive understanding of specific domains. Furthermore, in highly specialized fields, even human annotators face challenges, making it even more difficult for LLMs to grasp the correct logic and adhere to annotation standards.

Studies on faithfulness hallucinations in LMs focus on their uncertainty and overconfidence. \cite{diao2023active} introduces several metrics to characterize uncertainty, enabling the selection of the most uncertain questions for annotation through active prompting of LLMs. \cite{zhou2023navigating} investigates how epistemic markers—such as certainty, uncertainty, and evidentiality—affect LMs, concluding that LMs mimic observed language use rather than genuinely reflecting epistemic uncertainty. \cite{mundler2023self} proposes a novel prompting-based framework to detect and mitigate self-contradictions effectively.

\vspace{-0.06in}
\subsection{Knowledge Distillation}
\vspace{-0.03in}
Knowledge distillation transfers knowledge from a large teacher to a smaller student model. Leveraging the capabilities of LLMs in various machine learning tasks, many studies use LLMs as knowledge bases to enhance efficiency and improve responses. \cite{zhang2024efficient} introduces the Decision-Tree-of-Thought (DToT) prompting method, which boosts LLMs' detection performance and extracts high-quality rationales, improving overall performance and interpretability.

Studies also emphasize optimizing the interaction between teacher and student models during the distillation process. \cite{liu2024evolving} analyzes student model weaknesses and synthesizes labeled samples to help the teacher model address deficiencies effectively. \cite{senguptagood} proposes a collaborative approach with joint loss and curriculum learning for meta-teacher knowledge distillation. It creates a dynamic learning environment where the teacher model adapts its strategy based on the student's progress. However, in specialized domains where texts are often too professional to generate, we refine teaching templates using rationales to enhance interaction instead of developing new samples.

\begin{figure*}[t]
    \centering
    \includegraphics[width=\textwidth]{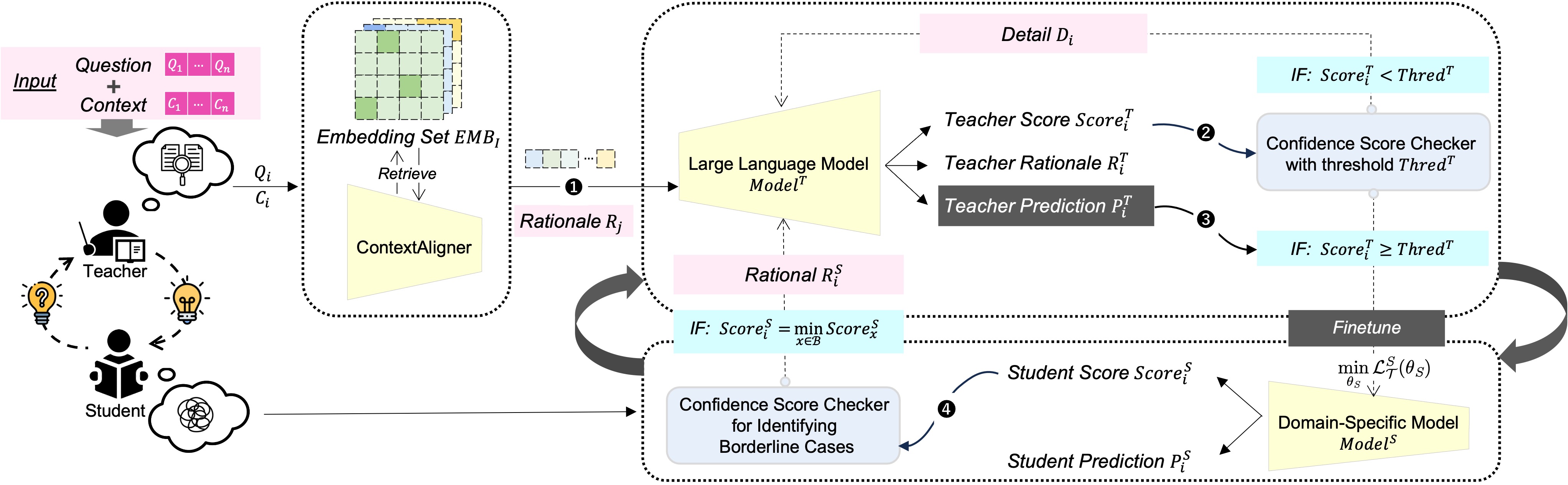}
    \caption{\texttt{DualChecker}: An Interactive Approach to Mitigate Hallucinations and Enhance Knowledge Distillation of LLMs. The process includes (1) \textit{ContextAligner}, which retrieves similar data and guides prompting using LLM-generated rationales; (2) the LLM then generates a confidence score, rationale, and prediction, which a \textit{Checker} evaluates against a threshold to determine the need for re-prompting; (3) if the confidence score meets the threshold, the prediction fine-tunes the student model, which outputs predictions and probabilities; (4) a second \textit{Checker} examines these probabilities, identifies the least confident case, generates a rationale, and updates the teaching template for subsequent prompting.}
    \label{fig:approach}
\end{figure*}

\vspace{-0.06in}
\subsection{Domain Adaptation}
\vspace{-0.03in}
Domain adaptation is a crucial application of LLMs, enhancing their ability to handle specialized tasks beyond general applications. For example, \cite{zhou2024lawgpt} develops LawGPT through legal-oriented pre-training and supervised fine-tuning. Similarly, \cite{li2024llava} leverages a biomedical figure-caption dataset, employs GPT-4 to generate self-instructed open-ended data, and trains a large vision-language model using a curriculum learning method. However, these approaches demand extensive pre-training or fine-tuning to achieve high performance.

\section{Approach}
Our proposed \texttt{DualChecker} addresses hallucinations in distilling LLMs, as illustrated in Figure \ref{fig:approach}. For a given task $\mathcal{T} \in \mathbb{T}$, with a question $Q_i$ and context $C_i$, the ContextAligner module within \texttt{DualChecker} first retrieves similar cases from the embedding set $EMB_I$, where $N=\{1,\dots,n\}$ is the set of all indices, $i$ is the current index and  $I = N \setminus \{i\}$. Then, a rationale $R_j$ is generated using LLMs for each similar case, where $j \in I$. Guided by these cases, the teacher model $Model^T$ produces a confidence score $Score_i^T$, rationale $R_i^T$, and prediction $P_i^T$. Next, a checker compares $Score_i^T$ against a threshold $Thred^T$. If the score is below the threshold, additional detail $D_i$ is incorporated for re-prompting; otherwise, $P_i^T$ is added to the batch $\mathcal{B}$ for fine-tuning the student model $Model^S$ with the objective function $\min_{\theta_S} \mathcal{L}_{\mathcal{T}}^S(\theta_S)$. The student model then outputs a probability score $Score_i^S$ and prediction $P_i^S$. Finally, a checker identifies the case in $\mathcal{B}$ with the lowest probability score. This borderline case, along with its LLM-generated rationale, is used by $Model^T$ to refine its teaching templates for the next prompting.

\begin{table*}[t]
\centering
\renewcommand{\arraystretch}{1.5}
\begin{adjustbox}{width=\textwidth}
\begin{tabular}{ccccc}
\hline
\textbf{Path Type} & \textbf{Description}\\ \hline \hline
\multicolumn{1}{c|}{0} & \multicolumn{1}{c}{Energy efficiency and consumption reduction - Content related to reducing all forms of energy consumption and improving efficiency} \\ \hline
\multicolumn{1}{c|}{1} & \multicolumn{1}{c}{Renewable energy and emission reduction - Content related to promoting the use of renewable energy and reducing emissions and greenhouse gases} \\ \hline
\multicolumn{1}{c|}{2} & \multicolumn{1}{c}{Waste management and recycling - Content related to waste reduction, improving recycling efficiency, and resource circulation} \\ \hline
\multicolumn{1}{c|}{3} & \multicolumn{1}{c}{Product development and technological innovation - Content related to developing new technologies and improving the durability and safety of products} \\ \hline
\end{tabular}
\end{adjustbox}
\caption{Path Classification Type.}\label{tbl:path_type}
\end{table*}

\vspace{-0.05in}
\subsection{Task Definition}
Our green innovation dataset comprises three core tasks: green innovation identification $\mathcal{T}^{cls}$, technological causality extraction $\mathcal{T}^{ce}$, and the identification of environmental impact pathways $\mathcal{T}^{path}$. The first task is a binary classification, the second involves token-level classification, and the third is a multi-class classification.

For a given context $C_i$, the question $Q_i$ for $\mathcal{T}^{cls}$ is: ``Does this text belong to the green innovation category?'' The answer is binary, either ``no'' (0) or ``yes'' (1). For $\mathcal{T}^{ce}$, $Q_i$ could be: ``Given the following text related to green innovation, extract the technical expression and the environmental effect it is expected to achieve.'' This task focuses on identifying causal relationships between the technical expression and its environmental impact, hence the term causality extraction. Finally, for $\mathcal{T}^{path}$, $Q_i$ asks: ``Given the following text related to green innovation, classify it into one of four categories.'' The classes for $\mathcal{T}^{path}$ are listed in Table \ref{tbl:path_type}, with the target answer ranging from 0 to 3, indicating the class index.

\vspace{-0.05in}
\subsection{ContextAligner} The ContextAligner module enhances alignment with human labeling standards, inspired by the RAG framework \cite{lewis2020retrieval} and preliminary experiments. Recognizing that LMs struggle with domain-specific tasks due to input length limitations, this module uses an existing labeled dataset to bridge the knowledge gap in few-shot in-context learning, instead of sourcing external contexts.
The core functionality generates an embedding vector $EMB_I$ for each context $C_i$. It computes the cosine similarity between context embeddings to identify the top $K$ most similar contexts. Specifically, each cosine similarity between $EMB_i$ and $EMB_{j \in I}$ is calculated as:
\begin{equation}
Similarity(EMB_i, EMB_j) = \frac{EMB_i^\top EMB_j}{\|EMB_i\| \|EMB_j\|}
\label{eq:similarity}
\end{equation}

For each retrieved-context $C_j$, the model uses LLMs to generate a rationale $R_j \sim \mathrm{LLM}(\:\cdot \mid C_j, L_j)$, aimed at improving the reasoning process. Here, $L_j$ represents the label of $C_j$. The aggregated data $(C_j, L_j, R_j)$ guides the teacher model, thereby facilitating more effective reasoning and decision-making.
\vspace{-0.05in}

\subsection{Teacher Confidence Checker}
The interactive checker systems function as a confidence checker for $Model^T$ and a borderline case identifier for $Model^S$. For the confidence checker, the teacher LLM first outputs a confidence score \(Score_i^T\), a rationale \(R_i^T\), and a prediction \(P_i^T\) from its reply: $Reply^T \sim \mathrm{LLM}(\:\cdot \mid C_j, L_j, R_j)$. Specifically, the confidence score \(Score_i^T\) is constrained to a range between 0\% and 100\%, with the additional instruction: ``Please output a confidence score as a percentage.'' The confidence checker assists in enhancing the consistency of the teacher model. The $Score_i^T$ is compared against a predefined teacher threshold $Thred^T$ to determine whether more details are needed to re-prompt. We assume that both black-box and white-box LLMs can generate explicit confidence scores during inference, and the checker system identifies model confidence as:

\begin{scriptsize}
\begin{equation}
\text{Checker}(Score_i^T) = \begin{cases}
\text{Unconfident,} & \text{if } Score_i^T \in [0\%, \text{Thred}^T) \\
\text{Confident,} & \text{if } Score_i^T \in [\text{Thred}^T, 100\%]
\end{cases}
\end{equation}
\end{scriptsize}

If $Score_i^T$ is identified as ``Unconfident,'' details of contexts $D_i$ are added to re-prompt the teacher model and regenerate the reply. Due to resource limitations, we set the re-prompting to occur only once. If identified as ``Confident,'' the teacher prediction $P_i^T$ is added to the batch for fine-tuning the student model.

\subsection{Student Fine-tuning}

Given the token representation of an input sequence $\mathit{Token} = \{\mathit{Token}_1,\dots,\mathit{Token}_t\}$, where $t$ is the length of the sequence and $k \in \{1, \dots, t\}$ represents the position of a token, the prediction probability for the classification tasks $\mathcal{T}^{cls}$ (binary), $\mathcal{T}^{path}$ (multi-class), and $\mathcal{T}^{ce}$ (causality extraction) is defined as:

\begin{equation}
p(c \mid \mathit{Token}) = \frac{\exp(\mathbf{w}_c^\top \mathbf{h})}{\sum_{m \in \mathcal{M}} \exp(\mathbf{w}_m^\top \mathbf{h})},
\end{equation}

where $\mathcal{M}$ is \{0, 1\} for $\mathcal{T}^{cls}$, \{0, 1, 2, 3\} for $\mathcal{T}^{path}$, and \{0, 1, 2, 3, 4\} for $\mathcal{T}^{ce}$. Here, $c \in \mathcal{M}$ is the class label, $\mathit{Token}$ represents the relevant token's representation (e.g., $\mathit{Token}_{cls}$ for [CLS] in sequence classification or $\mathit{Token}_k$ for the k-th token in token classification), $\mathbf{h}$ is the output of the last hidden layer corresponding to the relevant token (e.g., $\mathbf{h}_{cls}$ or $\mathbf{h}_k$), and $\mathbf{w}_c$ are trainable parameters. The summation is taken over all possible class labels $m \in \mathcal{M}$.

The fine-tuning loss $\mathcal{L}_\mathcal{T}^S$ for each task is defined as:

\begin{equation}
\mathcal{L}_\mathcal{T}^S = -\sum_{d \in \mathcal{D}} \log p(c^d \mid \mathit{Token}^d),
\end{equation}

where $d$ refers to each data point and $\mathcal{D}$ refers to the dataset corresponding to each task.

\subsection{Student Confidence Checker}
$Model^S$ outputs a prediction along with its corresponding probability score, denoted as $(Score_i^S, P_i^S) \sim \mathrm{Model}^S(\:\cdot \mid P_i^T)$. The borderline case identifier evaluates $Score_i^S$ when it is in batch $\mathcal{B}$, and determines whether it is the minimum score:

\begin{scriptsize}
\begin{equation}
\text{Checker}(Score_i^S) = \begin{cases}
\text{Unconfident,} & \text{if } Score_i^S = \min_{x \in \mathcal{B}} Score_x^S \\
\text{Confident,} & \text{otherwise}
\end{cases}
\end{equation}
\end{scriptsize}

If $Score_i^S$ is identified as ``Unconfident,'' the rationale $R_i^S \sim \mathrm{LLM}(\:\cdot \mid C_i^S, L_i^S)$, along with $C_i^S$ and $L_i^S$, is then applied to the teaching template of $Model^T$ for the next batch.

The process of \texttt{DualChecker}, which iteratively refines model predictions through context alignment and student-teacher interaction, is illustrated in Algorithm \ref{alg:dualchecker}.

\begin{algorithm}[H]
\caption{\texttt{DualChecker}}
\label{alg:dualchecker}
\begin{algorithmic}[1]
\REQUIRE Indices $N \in \{1,\dots,n\}$, Task $\mathcal{T} \in \mathbb{T}$, Question $Q_i$, Context $C_i$, Embedding Set $EMB_I$, Threshold $Thred^T$
\STATE \textbf{Initialization:} Retrieve similar cases from $EMB_I$ for $i$, $I = N \setminus \{i\}$

\STATE \textbf{Context Alignment:}
\FOR{each $j \in I$}
    \STATE Generate rationale $R_j$ using LLMs for each retrieved $C_j$
\ENDFOR

\STATE \textbf{Inference with Teacher Model $Model^T$:}
\STATE Produce $Score_i^T$, $R_i^T$, $P_i^T$
\IF{$Score_i^T < Thred^T$}
    \STATE Incorporate additional details $D_i$ for re-prompting
\ELSE
    \STATE Add $P_i^T$ to batch $\mathcal{B}$ for fine-tuning $Model^S$
\ENDIF

\STATE \textbf{Fine-tuning Student Model $Model^S$:}
\STATE Fine-tune $Model^S$ with $P_i^T$
\STATE Output $Score_i^S$, $P_i^S$
\STATE Identify borderline case $\arg \min_{x \in \mathcal{B}} Score_x^S$

\STATE \textbf{Student-Teacher Interaction:}
\IF{$Score_i^S == \min_{x \in \mathcal{B}} Score_x^S$}
    \STATE Generate rationale $R_i^S$ using LLMs
    \STATE Apply $R_i^S$, $C_i^S$, $L_i^S$ to update $Model^T$ templates
\ENDIF

\STATE \textbf{Output:} Updated $Model^S$
\end{algorithmic}
\end{algorithm}

\section{Experiments}

\begin{table*}[t]
\centering
\begin{tabular}{p{0.2\textwidth}p{0.07\textwidth}|p{0.05\textwidth}p{0.05\textwidth}p{0.05\textwidth}|p{0.05\textwidth}p{0.05\textwidth}p{0.05\textwidth}|p{0.05\textwidth}p{0.05\textwidth}p{0.05\textwidth}}
\hline
\noalign{\vskip 0.5pt}
\multirow{2}{*}{\textbf{Method}}
&\multirow{2}{*}{\textbf{N-Shot}} 
& \multicolumn{3}{c|}{$\mathcal{T}^{cls}$}
& \multicolumn{3}{c|}{$\mathcal{T}^{ce}$} 
& \multicolumn{3}{c}{$\mathcal{T}^{path}$} 
\\
~ & ~
& P & R & F1
& P & R & F1
& P & R & F1
\\\hline
\noalign{\vskip 0.5pt}
\multicolumn{1}{l}{GPT-3.5 Turbo}
&
&
&
&
&
&
&
&
&
&
\\
\multicolumn{1}{r}{-CoT}& 0-shot &
59.13&58.94&58.74&70.90&47.31&56.75&68.92&68.55&64.96\\
&5-shot & 
67.70&65.01&63.68&78.20&63.94&70.35&\textbf{74.34}&66.81&\underline{69.20}\\
\multicolumn{1}{r}{-EvoKD}& 0-shot & 
60.73&60.71&60.69&71.35&46.18&56.07&68.53&\underline{69.79}&65.78\\
&5-shot &
66.56&62.99&60.93&\textbf{79.17}&\underline{65.41}&\underline{71.63}&72.65&65.07&67.58\\
\multicolumn{1}{r}{-DToT}& 0-shot & 
60.39&59.36&58.37&70.76&49.38&58.17&65.72&64.06&59.85\\
&5-shot & 
\underline{68.64}&\underline{67.96}&\underline{67.68}&\underline{78.58}&65.25&71.30&\underline{72.70}&61.77&64.88\\
\hline
\noalign{\vskip 0.5pt}
\multicolumn{1}{r}{-DualChecker}& 0-shot &
57.49&55.37&52.01&59.78&43.10&50.09&68.56&63.76&60.96\\
&5-shot &
\textbf{84.85}&\textbf{84.83}&\textbf{84.83}&75.16&\textbf{73.27}&\textbf{74.20}&\underline{72.70}&\textbf{73.56}&\textbf{72.52}\\\hline
\noalign{\vskip 0.5pt}
\multicolumn{1}{l}{Llama 2}
&
&
&
&
&
&
&
&
&
&
\\
\multicolumn{1}{r}{-CoT}& 0-shot & \underline{61.13}&\underline{59.37}&\underline{57.51}&\textbf{80.32}&24.52&37.56&36.82&32.35&31.19\\
&5-shot & 
56.07&54.36&50.94&76.32&19.27&30.77&24.18&24.22&19.25\\
\multicolumn{1}{r}{-EvoKD}& 0-shot & 52.65&52.44&51.51&\underline{79.93}&15.83&26.43&39.81&\underline{39.78}&\underline{39.17}\\
&5-shot & 54.18&52.67&48.05&74.05&57.38&64.66&40.19&30.37&21.38\\
\multicolumn{1}{r}{-DToT}& 0-shot & 53.20&51.89&46.51&47.98&\textbf{64.92}&55.18&39.13&39.09&34.74\\
&5-shot &
58.73&54.00&46.87&75.54&62.97&\textbf{68.68}&\underline{42.66}&30.74&20.79\\
\hline
\noalign{\vskip 0.5pt}
\multicolumn{1}{r}{-DualChecker}& 0-shot & 
58.47&54.97&50.01&52.16&61.05&56.26&36.03&33.25&28.61\\
&5-shot & 
\textbf{67.74}&\textbf{65.11}&\textbf{63.38}&67.92&\underline{64.00}&\underline{65.90}&\textbf{47.09}&\textbf{43.88}&\textbf{43.65}\\
\Xhline{3\arrayrulewidth}

\end{tabular}
\caption{Comparison of teacher model performance between \texttt{DualChecker} and other baselines. Precision (P), Recall (R), and F1 Score (F1) are presented as percentages (\%). Bold text highlights the best results, and underlined text indicates the second-best.}
\vspace{-0.1in}
\label{tbl:teacher}
\end{table*}

\subsection{Datasets}
We evaluate \texttt{DualChecker} using a specialized textual dataset focused on green innovation, annotated by three domain experts. The dataset comprises 10,820 entries of green patent texts sourced from the Japan Patent Office (JPO), covering the period from 2006 to 2022. This dataset supports three primary tasks:

\subsubsection{Green Innovation Identification.}
This task aims to classify whether a given text pertains to green innovation. We achieved this by aligning all patent texts with the IPC codes listed in the Green Transformation Technologies Inventory (GXTI) standard, as provided by the JPO.\footnote{\url{https://www.jpo.go.jp/e/resources/statistics/gxti.html}} The task is structured as a binary classification problem with an equal distribution of 5,410 negative and 5,410 positive samples.

\subsubsection{Technological Causality Extraction.}
This task focuses on extracting technological phrases along with their corresponding environmental impact phrases. The dataset for this task consists of 1,000 labeled samples. The cause phrases have an average length of 42 words, with a maximum of 128 and a minimum of 4. The effect phrases exhibit average, maximum, and minimum lengths of 35, 142, and 7 words, respectively.

\subsubsection{Environmental Path Identification.}
This task aims to determine the environmental impact of green technologies, utilizing 1,000 labeled samples. Each sample is annotated based on the environmental pathways generated by GPT-4 Turbo and categorized into four distinct classes. These classes, ranging from 0 to 3, are summarized in Table \ref{tbl:path_type}. The distribution of samples across these classes is as follows: 374 samples for Class 0, 166 for Class 1, 90 for Class 2, and 72 for Class 3.

\vspace{-0.03in}
\subsection{Baselines}
\subsubsection{Backbones.}
We use both black-box and white-box LLMs as teacher models: GPT-3.5 Turbo and ELYZA-japanese-Llama-2-13b-instruct \cite{elyzallama2023}. For the student model, we select japanese-roberta-base \cite{sawada2024release} and train a patent-specific RoBERTa model from scratch on 3 million open-source patents. Additionally, we fine-tune a patent-specific Sentence-RoBERTa model to generate textual embeddings for ContextAligner. We also release these models to enhance the reproducibility of our study. Please see Appendix \ref{appendix:model} for more details on these models.

\subsubsection{CoT.}
We apply the chain-of-thought (CoT) strategy \cite{wei2022chain} to benchmark the reasoning capabilities of our \texttt{DualChecker}. CoT guides LLMs through intermediate reasoning steps, enabling them to tackle complex tasks more effectively.

\subsubsection{DToT.}
The Decision-Tree-of-Thought (DToT) \cite{zhang2024efficient} is a reflection strategy that uses confidence scores and rationales to improve the reliability of LLM outputs. We employ DToT to benchmark the effectiveness of mitigating teacher models' hallucinations compared to our method.

\subsubsection{EvoKD.}
Evolving Knowledge Distillation (EvoKD) \cite{liu2024evolving} enhances the data augmentation process using LLMs to improve interaction during knowledge distillation. We select EvoKD to compare the robustness of its interaction process with that of our method.

\begin{table*}[t]
\centering
\begin{tabular}{p{0.3\textwidth}|p{0.05\textwidth}p{0.05\textwidth}p{0.05\textwidth}|p{0.05\textwidth}p{0.05\textwidth}p{0.05\textwidth}|p{0.05\textwidth}p{0.05\textwidth}p{0.05\textwidth}}
\hline
\noalign{\vskip 0.5pt}
\multirow{1}{*}{\textbf{Method}} 
& \multicolumn{3}{c|}{$\mathcal{T}^{cls}$}
& \multicolumn{3}{c|}{$\mathcal{T}^{ce}$} 
& \multicolumn{3}{c}{$\mathcal{T}^{path}$} 
\\
~
& P & R & F1
& P & R & F1
& P & R & F1
\\\hline
\noalign{\vskip 0.5pt}
\multicolumn{1}{l|}{Patent RoBERTa}
&
&
&
&
&
&
&
&
&
\\
\multicolumn{1}{r|}{\textit{-human}}
&\textit{91.45}&\textit{91.45}&\textit{91.45}&\textit{49.30}&\textit{47.90}&\textit{48.57}&\textit{12.84}&\textit{82.74}&\textit{79.94}\\
\multicolumn{1}{r|}{-CoT (GPT-3.5 Turbo)}&
70.24&64.80&62.04&\textbf{49.71}&\underline{41.81}&\underline{44.46}&70.66&63.80&\underline{66.46}\\
\multicolumn{1}{r|}{-EvoKD (GPT-3.5 Turbo)}&
71.92&56.82&47.53&48.76&41.35&43.96&68.71&\underline{64.03}&65.93\\
\multicolumn{1}{r|}{-DToT (GPT-3.5 Turbo)}&
\underline{76.14}&\underline{75.69}&\underline{75.52}&\underline{49.29}&41.19&43.99&\underline{71.40}&60.96&64.34\\
\hline
\noalign{\vskip 0.5pt}
\multicolumn{1}{r|}{-DualChecker (GPT-3.5 Turbo)}
& \textbf{85.48} & \textbf{85.47} & \textbf{85.44} & 48.78 & \textbf{44.03} & \textbf{45.95} & \textbf{72.86} & \textbf{75.25} & \textbf{73.78}
\\\hline
\noalign{\vskip 0.5pt}
& & & & & & & & &\\
\multicolumn{1}{r|}{-CoT (Llama 2)}
&\underline{56.81}&\underline{54.83}&\underline{51.10}&43.95&40.19&41.90&9.26&22.50&13.13\\
\multicolumn{1}{r|}{-EvoKD (Llama 2)}
&53.26&51.74&45.12&\underline{45.72}&\textbf{42.06}&\textbf{43.80}&8.75&25.00&12.96\\
\multicolumn{1}{r|}{-DToT (Llama 2)}&
50.19&50.08&41.25&\textbf{46.27}&\underline{40.50}&\underline{42.67}&\underline{14.00}&\underline{30.44}&\underline{18.67}\\
\hline
\noalign{\vskip 0.5pt}
\multicolumn{1}{r|}{-DualChecker (Llama 2)}
&\textbf{64.18}&\textbf{64.16}&\textbf{64.16}&41.88&39.77&40.64&\textbf{30.50}&\textbf{33.35}&\textbf{27.03}\\\hline
\noalign{\vskip 0.5pt}
\multicolumn{1}{l|}{RoBERTa}
&
&
&
&
&
&
&
&
&
\\
\multicolumn{1}{r|}{\textit{-human}}&
\textit{89.23}&\textit{89.23}&\textit{89.23}&\textit{49.19}&\textit{44.47}&\textit{46.13}&\textit{76.18}&\textit{78.65}&\textit{76.94}\\
\multicolumn{1}{r|}{-CoT (GPT-3.5 Turbo)}&
79.21&70.05&67.23&47.44&\underline{43.75}&\underline{45.38}&\underline{76.66}&\underline{73.06}&\underline{74.60}\\
\multicolumn{1}{r|}{-EvoKD (GPT-3.5 Turbo)}&
24.68&50.00&33.04&\underline{48.05}&42.13&44.30&71.22&64.19&66.84\\
\multicolumn{1}{r|}{-DToT (GPT-3.5 Turbo)}&
\underline{81.24}&\underline{80.77}&\underline{80.62}&\textbf{48.68}&41.85&44.32&\textbf{77.90}&68.12&71.31\\
\hline
\noalign{\vskip 0.5pt}
\multicolumn{1}{r|}{-DualChecker(GPT-3.5 Turbo)}
& 
\textbf{85.26}&\textbf{85.16}&\textbf{85.11}&47.88&\textbf{43.92}&\textbf{45.71}&74.83&\textbf{79.58}&\textbf{76.43}\\
\hline
\noalign{\vskip 0.5pt}
& & & & & & & & &\\
\multicolumn{1}{r|}{-CoT (Llama 2)}
&\underline{55.33}&\underline{53.81}&\underline{50.09}&42.73&\underline{41.67}&42.11&8.75&25.00&12.96\\
\multicolumn{1}{r|}{-EvoKD (Llama 2)}
&54.40&52.46&46.36&44.39&\textbf{42.73}&\textbf{43.48}&8.75&25.00&12.96\\
\multicolumn{1}{r|}{-DToT (Llama 2)}&
47.64&49.46&37.13&\underline{44.44}&40.47&42.21&\underline{16.40}&\textbf{38.45}&\underline{22.15}\\
\hline
\noalign{\vskip 0.5pt}
\multicolumn{1}{r|}{-DualChecker(Llama 2)}
& 
\textbf{68.21}&\textbf{64.17}&\textbf{61.86}&\textbf{45.75}&40.36&\underline{42.32}&\textbf{29.18}&\underline{36.16}&\textbf{30.78}\\
\noalign{\vskip 0.5pt}
\Xhline{3\arrayrulewidth}

\end{tabular}
\caption{Comparison of student model performance between \texttt{DualChecker} and other baselines. Precision (P), Recall (R), and F1 Score (F1) are presented as percentages (\%). Bold text highlights the best results, and underlined text indicates the second-best.}
\vspace{-0.1in}
\label{tbl:student}
\end{table*}

\vspace{-0.03in}

\subsection{Implementations}
To ensure a fair comparison, we maintain the original settings of the baseline models. The parameters are as follows: 1) a batch size of 8 for distillation, 64 for fine-tuning \(\mathcal{T}^{cls}\), and 8 for \(\mathcal{T}^{path}\) and \(\mathcal{T}^{ce}\); 2) a training ratio of 0.8; 3) a learning rate of 2e-5; and 4) teacher model confidence thresholds set at 0.85 for GPT-3.5 Turbo and 0.75 for Llama 2. The experiments were conducted on 5 NVIDIA RTX A6000 GPUs. See Appendices \ref{appendix:para} and \ref{appendix:template} for details.

\subsection{Results}
Through experiments on benchmark datasets, we aim to address the following research questions:

\begin{figure}[t]
\centering
\includegraphics[width=\linewidth]{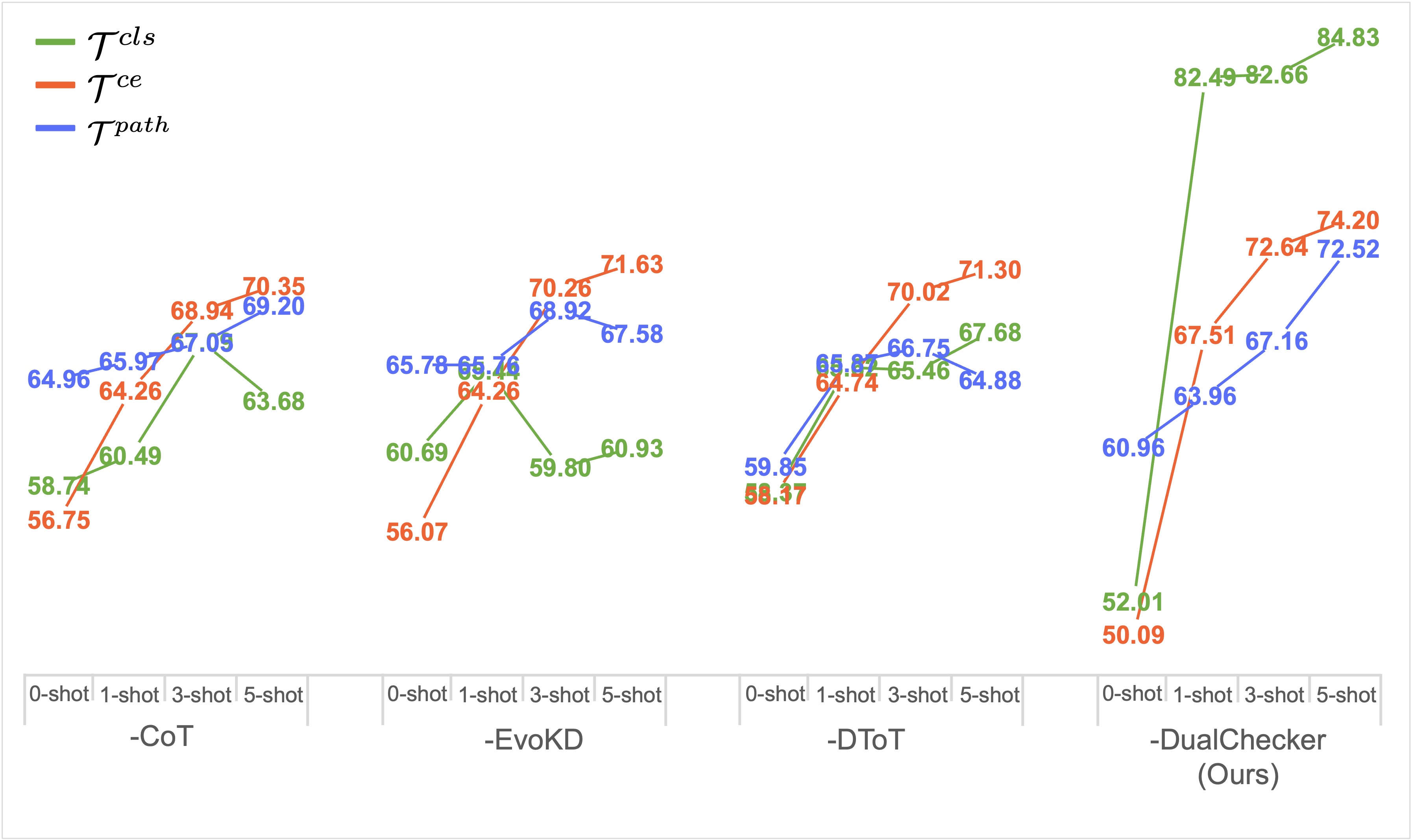}
\vspace{-0.08in}
\caption{Results with Varying Numbers of Shots using GPT-3.5 Turbo.}
\label{fig:teacher_n_shot}
\vspace{-0.17in}
\end{figure}
\subsubsection{Q1: Can \texttt{DualChecker} effectively mitigate the hallucinations of teacher LLMs?}
Table \ref{tbl:teacher} compares \texttt{DualChecker} with other baselines across both black-box and white-box LLMs in 0-shot and 5-shot settings. Although our method is specifically tailored for few-shot in-context learning, 0-shot results are included for reference. \texttt{DualChecker} consistently outperforms the baselines, achieving substantial F1 score improvements—up to \textbf{17\%}\footnote{Gain is defined as the difference in F1 scores between \texttt{DualChecker} and other baselines using different backbone models. The 0-shot scenario of \texttt{DualChecker} is excluded as our method is based on few-shot learning.} in $\mathcal{T}^{cls}$ with GPT-3.5 Turbo (5-shot), \textbf{3\%} in $\mathcal{T}^{ce}$ with the same model, and \textbf{4\%} in $\mathcal{T}^{path}$ with Llama 2 (5-shot). These significant gains underscore \texttt{DualChecker}'s robust ability to reduce hallucinations across diverse tasks and models, particularly in scenarios requiring high factual accuracy and specialized domain adaptation.

Notably, the comparison between 0-shot and 5-shot settings reveals that \texttt{DualChecker} enhances F1 scores by 10\% to 30\%, clearly demonstrating its effectiveness in addressing hallucination issues in reasoning tasks. The pronounced gains observed with GPT-3.5 Turbo highlight \texttt{DualChecker}’s exceptional impact when applied to advanced LLMs. Moreover, the significant improvement in $\mathcal{T}^{cls}$ illustrates the model's adeptness at leveraging minimal training data to achieve accurate classification, even in training-free scenarios. This reinforces \texttt{DualChecker} as a powerful tool for enhancing the reliability and accuracy of teacher models in complex settings.

\vspace{-0.0in}
\subsubsection{Q2: How does the number of shots influence the performance of \texttt{DualChecker}?}
We evaluated \texttt{DualChecker} with varying numbers of shots using GPT-3.5 Turbo to assess its robustness across a range of tasks. As shown in Figure \ref{fig:teacher_n_shot}, the performance of \texttt{DualChecker} significantly improves as the number of shots increases, which aligns with its design for few-shot settings. Notably, the most substantial performance gain is observed after 3-shot, highlighting the model's ability to adapt and improve with limited labeled data. Given that only the ContextAligner component is active in the 1-shot scenario, these results also emphasize the critical role of the interaction process between models, which ensures continuous improvement as more shots are added. This adaptability and responsiveness to additional shots demonstrate \texttt{DualChecker}'s strong potential in dynamic and evolving learning environments.

\vspace{-0.0in}
\subsubsection{Q3: Can \texttt{DualChecker} enhance the performance of student models using the predictions of teacher models?}

Table \ref{tbl:student} compares the performance of student models fine-tuned with different labels. The \textit{human} labels refer to ground truth annotations provided by domain experts, while the predictions from GPT-3.5 Turbo and Llama 2 are based on 5-shot scenarios using various methods. In the experiment with Patent RoBERTa, the improvement gains with GPT-predictions are \textbf{9.9\%}, \textbf{1.5\%}, and \textbf{7.3\%} in $\mathcal{T}^{cls}$, $\mathcal{T}^{ce}$, and $\mathcal{T}^{path}$, respectively. At the same time, the Llama-predictions yield gains of \textbf{13.1\%} and \textbf{8.4\%} in $\mathcal{T}^{cls}$ and $\mathcal{T}^{path}$. These results underscore the effectiveness of \texttt{DualChecker} in boosting student model performance by leveraging teacher model predictions.

The significant gains observed in $\mathcal{T}^{cls}$ and $\mathcal{T}^{path}$ demonstrate that \texttt{DualChecker} is particularly effective in tasks centered on sequence classification. However, in $\mathcal{T}^{ce}$, the improvement is more modest, likely due to the task's inherent specificity and the challenges of causality extraction. The subjectivity in the annotation rules for causality extraction makes it harder to achieve substantial gains by adding similar cases, rationales, and reflections, compared to more straightforward classification tasks. This highlights the nuanced challenges different tasks present and identifies where \texttt{DualChecker} can most effectively drive performance enhancements.

\vspace{-0.0in}
\subsubsection{Q4: Is \texttt{DualChecker} a robust framework for distilling LLMs in a challenging domain?}

To assess the robustness of \texttt{DualChecker} with different student models, we also experimented with a RoBERTa model trained on a general corpus as shown in Table \ref{tbl:student}. Since this model lacks comprehensive domain knowledge, its performance is relatively lower. Despite this, we still observed significant improvement gains with GPT-predictions: \textbf{4.5\%}, \textbf{0.3\%}, and \textbf{1.8\%} in $\mathcal{T}^{cls}$, $\mathcal{T}^{ce}$, and $\mathcal{T}^{path}$, respectively. At the same time, the Llama-predictions yielded gains of \textbf{11.8\%} and \textbf{8.6\%} in $\mathcal{T}^{cls}$ and $\mathcal{T}^{path}$. Additionally, when comparing the scores of \texttt{DualChecker} with \textit{human} labels, our method approaches the gold standard closely, within a 5\% difference, whereas other baselines exhibit more than a 10\% difference. This comparable performance indicates that \texttt{DualChecker} can effectively leverage existing knowledge to guide LLMs and student models robustly, even when these models lack specific domain knowledge.

\vspace{-0.0in}
\subsubsection{Q5: Which component of \texttt{DualChecker} contributes the most to improvement?}
To understand the contribution of each \texttt{DualChecker} component, we conduct an ablation study using GPT-3.5 Turbo predictions under a 5-shot setting (see Table \ref{tbl:ablation}). ``-w/ContextAligner'' refers to using only the ContextAligner, ``-w/Teacher'' includes both the ContextAligner and the checker system in teacher model reasoning, and ``-w/Student'' consists of both in student model fine-tuning.

The results reveal that ContextAligner alone provides a substantial improvement, highlighting its effectiveness in aligning LLMs with human annotation standards. Surprisingly, ``-w/Teacher'' and ``-w/Student'' perform worse than ContextAligner alone. This may occur because teacher models, without feedback from student models, lose direction for further refinement. Additionally, teacher LLMs may become ``overconfident'' \cite{santurkar2023whose}, leading to overly optimistic and biased predictions. The complete \texttt{DualChecker} system outperforms all others by integrating improvements for both teacher and student models.
\begin{table}[t]
\centering
\begin{tabular}{p{0.2\textwidth}|p{0.05\textwidth}|p{0.05\textwidth}|p{0.05\textwidth}}
\hline
\noalign{\vskip 0.5pt}
\multirow{1}{*}{\textbf{Method}} 
& \multicolumn{1}{c|}{$\mathcal{T}^{cls}$}
& \multicolumn{1}{c|}{$\mathcal{T}^{ce}$} 
& \multicolumn{1}{c}{$\mathcal{T}^{path}$} 
\\
~
& F1
& F1
& F1
\\\hline
\noalign{\vskip 0.5pt}
\multicolumn{1}{l|}{Dualchecker}
&
&
&
\\
\multicolumn{1}{r|}{-w/ContextAligner}&
84.32&74.12&67.70\\
\multicolumn{1}{r|}{-w/Teacher}&
84.01&73.90&67.23\\
\multicolumn{1}{r|}{-w/Student}&
84.31&73.06&62.69\\
\hline
\noalign{\vskip 0.5pt}
\multicolumn{1}{r|}{-all}
& 
84.83&74.20&72.52\\
\noalign{\vskip 0.5pt}
\Xhline{3\arrayrulewidth}
\end{tabular}
\caption{Ablation study of \texttt{DualChecker} using GPT-3.5 Turbo in the 5-shot setting.}
\vspace{-0.1in}
\label{tbl:ablation}
\end{table}

\section{Conclusions}
In this study, we introduce \texttt{DualChecker}, a novel framework that effectively mitigates hallucinations in large language models during knowledge distillation. Unlike traditional approaches that rely on external knowledge or extensive training, \texttt{DualChecker} employs ContextAligner and an interactive checker system to align model outputs with human standards, ensuring accuracy, consistency, and reliability. Our extensive experiments in the challenging green innovation domain demonstrate that \texttt{DualChecker} significantly outperforms existing methods, achieving notable F1 score improvements for both teacher and student models. The framework's adaptability across black and white-box models further underscores its robustness and versatility. By open-sourcing our datasets, models, and code, we aim to foster further research and accelerate the development of more reliable and effective AI systems in complex, real-world applications.

\appendix

\section{Model Details}
\label{appendix:model}
We employ GPT-3.5 Turbo with its default configuration (e.g., temperature set to 1). Detailed specifications for other pre-trained models are provided in Table \ref{tbl:details}, sourced from HuggingFace. For our patent-specific RoBERTa model, we leverage 3 million unlabeled patent texts, configuring it with a batch size of 1,904 and training it for 1 million steps. Additionally, we have developed a patent-specific Sentence-RoBERTa model by fine-tuning the base patent RoBERTa model using the JSNLI dataset \footnote{\url{https://huggingface.co/datasets/shunk031/jsnli}}, a Japanese adaptation of the SNLI dataset \footnote{\url{https://nlp.stanford.edu/projects/snli/}}, comprising 533,005 training samples and 3,916 validation samples. Both models will be publicly available for broader use.
\begin{table}[H]
\centering
\begin{adjustbox}{width=\linewidth}
\begin{tabular}{ccc}
\hline

\textbf{Model}                            & \textbf{HuggingFace Key}                              & \textbf{Model Size}  \\ \hline \hline
\multicolumn{1}{c|}{Llama 2}
& \multicolumn{1}{c|}{elyza/ELYZA-japanese-Llama-2-13b-instruct} 
& 
26.03GB
\\
\hline
\multicolumn{1}{c|}{RoBERTa}
&
\multicolumn{1}{c|}{rinna/japanese-roberta-base}
&
443MB
\\
\cline{1-3}
\end{tabular}
\end{adjustbox}
\caption{Details of pre-trained models.}\label{tbl:details}
\end{table}

\section{Experimental Settings}
\label{appendix:para}
In the few-shot in-context learning of LLMs, we initially select examples from the test data and incorporate the rationale generated by GPT-4 Turbo along with the confidence scores labeled by experts. Our preliminary experiment found that LLMs exhibit overconfidence, ranging from 90\% to 100\%. To mitigate this, we set the confidence scores in the templates to range from 60\% to 90\%. Meanwhile, the maximum similar contexts retrieved from ContextAligner can be expressed as:

\begin{equation}
\text{Max Similar Contexts} = \begin{cases}
0, & \text{if } n = 0 \\
1, & \text{if } n = 1, 2, 3 \\
\left\lceil \frac{n-1}{2} \right\rceil, & \text{if } n > 3
\end{cases}
\end{equation}

where \( n \) is the number of shots. This helps reduce the bias in instruction. The borderline cases identified by the student model will replace the examples in the teaching template for the next round of reasoning.

\section{Template Types}
\label{appendix:template}

The teaching templates for each task are set as shown in Table \ref{tbl:template}:

\begin{table}[H]
\centering
\begin{adjustbox}{width=\linewidth}
\begin{tabular}{c|p{0.85\linewidth}}
\hline
\textbf{Task} & \textbf{Description} \\ 
\hline \hline
\multicolumn{1}{c|}{$\mathcal{T}^{cls}$} & Determine if the following text belongs to the Green Innovation category. Answer with 'yes' or 'no', and rate your confidence on a scale of 100 points. Return the answer, confidence, and rationale in the following JSON format: \{"Answer": Answer, "Confidence": Confidence, "Rationale": Rationale\}. \\ 
\hline
\multicolumn{1}{c|}{$\mathcal{T}^{ce}$} & Identify the technology and the ultimate environmental effect within the following sentence related to Green Innovation. Extract both the technology and the environmental effect, and rate your confidence on a scale of 100 points. Return the technology, environmental effect, confidence, and rationale in the following JSON format: \{"Technology": Technology, "Environmental Effect": Environmental Effect, "Confidence": Confidence, "Rationale": Rationale\}. \\ 
\hline
\multicolumn{1}{c|}{$\mathcal{T}^{path}$} & Classify the environmental issue that the technology in the following sentence related to Green Innovation can ultimately resolve, using the labels (0,1,2,3): 0: Energy efficiency and consumption reduction - Content related to reducing all forms of energy consumption and improving efficiency, 1: Renewable energy and emission reduction - Content related to promoting the use of renewable energy and reducing emissions and greenhouse gases, 2: Waste management and recycling - Content related to waste reduction, improving recycling efficiency, and resource circulation, 3: Product development and technological innovation - Content related to developing new technologies and improving the durability and safety of products. Select one label from (0,1,2,3), and return the label and rationale in the following JSON format: \{"Label": Label, "Rationale": Rationale\}. \\
\hline
\end{tabular}
\end{adjustbox}
\caption{Templates for different tasks.}
\label{tbl:template}
\end{table}
\newpage
\setcounter{secnumdepth}{0}
\bibliography{paper}
\end{document}